\documentclass[twoside]{article}
\usepackage[utf8]{inputenc}
\usepackage{amsmath}
\usepackage{amsfonts}
\usepackage{amssymb}
\usepackage{graphicx}
\usepackage{color}
\usepackage[normalem]{ulem}
\usepackage[left=2cm,right=2cm,top=2cm,bottom=2cm]{geometry}
\newenvironment{Proof}{\noindent{\sc Proof.}}{\qed}
\newtheorem{theorem}{Theorem}[section]
\newtheorem{lemma}{Lemma}[section]

\newtheorem{rem}{Remark}[section]

\newtheorem{uda}{Example}[section]
\newcommand{\qed}{\hfill$\Box$\par\medskip}

\def\bhag#1{\noindent
\setcounter{equation}{0}
\section{#1}
}

\def\RR{{\mathbb R}}

\def\ZZ{{\mathbb Z}}

\def\PPI{{{\rm I}\kern-1pt\Pi}}

\def\b #1;{{\bf #1}}
\def\x{{\bf x}}
\def\k{{\bf k}}
\def\y{{\bf y}}
\def\u{\mathbf{u}}
\def\w{{\bf w}}
\def\z{{\bf z}}

\def\e{\epsilon}

\def\t{\mathbf{t}}

\def\O{{\cal O}}

\def\C{{\mathcal C}}

\def\be{\begin{equation}}
\def\ee{\end{equation}}
\def\bea{\begin{eqnarray}}
\def\eea{\end{eqnarray}}
\def\eref#1{(\ref{#1})}
\def\disp{\displaystyle}

\def\donchitre#1#2{\vskip 6.5cm\noindent
\parbox[t]{1in}{\special{eps:#1.eps x=6.5cm y=5.5cm}}
\hbox to 7cm{}\parbox[t]{0.0cm}{\special{eps:#2.eps x=6.5cm y=5.5cm}}}

\def\XX{{\mathbb X}}

\def\bs#1{{\boldsymbol{#1}}}
\def\bsw{\bs{\omega}}

\title{Deep nets for local manifold learning}
\author{Charles K. Chui\thanks{Department of Statistics, Stanford  University, Stanford, CA 94305. The research of this author is supported by ARO Grant W911NF-15-1-0385.
\textsf{email:} ckchui@stanford.edu. }\ 
\  and
H.~N.~Mhaskar\thanks{Department of Mathematics, California Institute of Technology, Pasadena, CA 91125;
Institute of Mathematical Sciences, Claremont Graduate University, Claremont, CA 91711. The research of this author is supported in part by ARO Grant W911NF-15-1-0385.
\textsf{email:} hrushikesh.mhaskar@cgu.edu. }       }
\date{}
\begin{document}
\maketitle
\begin{abstract}
The problem of extending a function $f$ defined on a training data $\C$ on an
unknown manifold $\XX$ to the entire manifold and a tubular neighborhood of this manifold is considered in this paper.  For $\XX$ embedded in a high dimensional ambient Euclidean space $\RR^D$,  a deep learning algorithm is developed for finding a local coordinate system for the manifold \textbf{without eigen--decomposition}, which reduces the problem to the classical problem of function approximation on a low dimensional cube. Deep nets (or multilayered neural networks) are proposed to accomplish this approximation scheme by using the training data. Our methods do not involve such optimization techniques as back--propagation, while assuring optimal (a priori) error bounds on the output in terms of the number of derivatives of the target function. In addition, these methods are universal, in that they do not require a prior knowledge of the smoothness of the target function, but adjust the accuracy of approximation locally and automatically, depending only upon the local smoothness of the target function. Our ideas are easily extended to solve both the pre--image problem and the out--of--sample extension problem, with a priori bounds on the growth of the function thus extended. 
\end{abstract}

\bhag{Introduction}
Machine learning is an active sub--field of Computer Science on algorithmic development for learning and making predictions based on some given data, with a long list of applications that range from computational finance and advertisement, to information retrieval, to computer vision, to speech and handwriting recognition, and to structural healthcare and medical diagnosis. In terms of function approximation, the data for learning and prediction can be formulated as $\{(\x, f_\x)\}$, obtained with an unknown probability distribution. Examples include: the Boston housing problem (of predicting the median price $f_\x$ of a home based on some vector $\x$ of 13 other attributes \cite{vapnik2013nature}) and the floor market problem \cite{tiao1989model, chakraborty1992forecasting} (that deals with the indices of the wheat floor pricing in three major markets in the United States). For such problems, the objective is to predict the index $f_\x$ in the next month, say, based on a vector $\x$ of their values over the past few months. Other similar problems include the prediction of blood glucose level $f_\x$ of a patient based on a vector $\x$ of the previous few observed levels \cite{sergei, mnpdiabetes}, and the prediction of box office receipts ($f_\x$) on the date of release  of a movie in preparation, based on a vector $\x$ of the survey results about the movie \cite{sharda2002forecasting}.  It is pointed out in \cite{multilayer, mauropap, compbio} that all the pattern classification problems can also be viewed fruitfully as problems of function approximation. While it is an ongoing research to allow non--numeric input $\x$ (e.g., \cite{treepap}), we restrict our attention in this paper to the consideration of $\x\in\RR^D$, for some integer $D\ge 1$.

In the following discussion, the first component $\x$ is considered as input, while the second component $f_\x$ is considered the output of the underlying process. The central problem is to estimate the conditional expectation of $f_\x$ given $\x$. Various statistical techniques and theoretical advances in this direction are well--known (see, for example \cite{vapnik1998statistical}). In the context of neural and radial--basis--function networks, an explicit formulation of the input/output machines was pointed out in \cite{girosi1990networks,girosi1995regularization}. More recently, the nature of deep learning as an input/output process is formulated in the same way, as explained in \cite{lecun2015deep, poggio_deep_net_2015}. To complement the statistical perspective and understand  the theoretical capabilities of these processes, it is customary to think of the expected value of $f_\x$,  given $\x$ , as a function $f$ of $\x$. The question of empirical estimation in this context is to carry out the approximation of $f$ given samples $\{(\x, f(\x))\}_{\x\in\C}$, where $\C$ is a finite \emph{training data} set. In practice, because of the random nature of the data, it may be possible that there are several pairs of the form $(\x, f_\x)$ in the data for the same values of $\x$. In this case, a statistical scheme, such as some kind of averaging of $f_\x$ being the simplest one, can be used to obtain a desired value $f(\x)$ for the sample of $f$ at $\x$, $\x\in\C$. From this perspective, the problem of extending $f$ from the traning data set $\C$ to $\x$ not in $\C$ in machine  learning is called the \emph{generalization problem}.

We will  illustrate this general line of ideas by using neural networks as an example. To motivate this idea, let us first recall a theorem originating with Kolmogorov and Arnold \cite[Chapter~17, Theorem~1.1]{lorentz_advanced}. According to this theorem, there exist \emph{universal} Lipschitz continuous functions $\phi_1,\cdots,\phi_{2D+1}$ and \emph{universal} scalars $\lambda_1,\cdots,\lambda_D\in (0,1)$, for which every continuous function $f : [0,1]^D\to \RR$ can be written as 
\be\label{kolmtheo}
f(\x)=\sum_{j=1}^{2D+1}g\left(\sum_{k=1}^D \lambda_k\phi_j(x_k)\right), \qquad \x=(x_1,\cdots,x_D)\in [0,1]^D,
\ee
where $g$ is a continuous function that depends on $f$. In other words, for a given $f$, only one function $g$ has to be determind to give the representation formula \eref{kolmtheo} of $f$. 

A neural network, used as an input/output machine, consists of an input layer, one or more hidden layers, and an output layer. Each hidden layer consists of a number of neurons arranged according to the network architecture. Each of these neurons has a local memory and performs a simple non--linear computation upon its input. The input layer fans out the input $\x\in\RR^D$ to the neurons at the first hidden layer. The output layer typically takes a linear combination of the outputs of the neurons at the last hidden layer. 
The right hand side of \eref{kolmtheo} is a neural network with two hidden layers. The first contains $D$ neurons, where the $j$--th neuron computes the sum $\sum_{k=1}^D \lambda_k\phi_j(x_k)$. The next hidden layer contains $2D+1$ neurons each evaluating the function $g$ on the 
output of the $j$--th neuron in the first hidden layer. The output layers takes the sum of the results as indicated in \eref{kolmtheo}.

From a practical point of view, such a network is clearly hard to construct, since only the existence of the functions $\phi_j$ and $g$ is known, without a numerical procedure for computing these. In the early mathematical development of neural networks during the late 1980s and early 1990s, instead of finding these functions for the representation of a given continuous function $f$ in \eref{kolmtheo}, the interest was to study the existence and characterization of  \emph{universal} functions $\sigma :\RR\to\RR$, called \emph{activation functions} of the neural networks, such that each neuron evaluates the activation function upon an affine transform of its input, and the network is capable of approximating any desired real-valued continuous target function $f: K\to\RR$ arbitrarily closely on $K$, where $K\subset \RR^D$ is any compact set.

For example, a neural network with one hidden layer can be expressed as a function
\be\label{onehiddenlayer}
\mathcal{N}(\x)=\mathcal{N}_n(\{\w_k\}, \{a_k\}, \{b_k\};\x)= \sum_{k=1}^n a_k\sigma(\w_k\cdot \x +b_k), \qquad \x\in\RR^D.
\ee
Here, the hidden layer consists of $n$ neurons, each of which has a local memory. The local memory of the $k$--th neuron contains the \emph{weights} $\w_k\in \RR^D$, and the \emph{threshold} $b_k\in\RR$. Upon receiving the input $\x\in\RR^D$ from the input later, the $k$--th neuron evaluates $\sigma(\w_k\cdot\x+b_k)$ as its output, where $\sigma$ is a non--linear activation function. The output layer is just one circuit where the coefficients  $\{a_k\}$ are stored in a local memory, and the evaluates the linear combination as indicated in \eref{onehiddenlayer}.  Training of this network in order to learn a function $f$ on a compact subset $K\subset\RR^D$ to an accuracy of $\e>0$ involves finding the parameters $\{a_k\}$, $\{\w_k\}$, $\{b_k\}$ so that 
\be\label{universal appr} 
\max_{\x\in K}|f(\x)-\mathcal{N}(\x)|<\e.
\ee
The most popular technique for doing this is the so called back--propagation, which seeks to find these quantities by minimizing an error functional usually with some regularization parameters. 
We remark that  the number $n$ of \emph{neurons} in the approximant \eref{onehiddenlayer} must increase, if the tolerance $\e >0$ in the approximation
of the target function $f$ is required to be smaller.

From a theoretical perspective, the main attraction of neural networks with one hidden layer is their \emph{universal approximation property} as formulated in \eref{universal appr}, which overshadows the properties of their predecessors, namely: the perceptrons \cite{minsky1988perceptrons}. In particular, the question of finding sufficient conditions on the actvation function $\sigma$ that ensure the universal approximation property was investigated in great detail by many authors, with emphasis on the most popular \emph{sigmoidal function}, defined by the property $\sigma(t)\to 1$ for $t\to\infty$ and $\sigma(t)\to 0$ for $t\to -\infty$. For example,  Funahashi \cite{funahashi1989} applied some discretization of an integral formula from \cite{irie1988} to prove the universal approximation property for some sigmoidal function $\sigma $.  A similar theorem was proved by Hornik, Stinchcombe, White \cite{hornik1989} by using the Stone--Weierstrass theorem, and another by Cybenko \cite{cybenko1989} by applying the Hahn--Banach and Riesz Representation theorems.  A constructive proof via approximation by ridge functions  was given in  our paper \cite{chuili1992}, with algorithm for implementation presented in our follow-up work \cite{chui1993realization}. A complete characterization of which activation functions are allowed to achieve the universal approximation property was given later in \cite{mhasmich, leshnolinpinkus}. 

However, for neural networks with one hidden layer, one of the severe limitations to applying training algorithms based on optimization, such as back--propagation or those proposed in the book \cite{vapnik1998statistical} of Vapnik, is that it is neccessary to know the number of neurons in $\mathcal{N}$ in advance. 
Therefore, one major problem in the 1990s, known as the \emph{complexity problem}, was to estimate the number of neurons required to approximate a function to a desired accuracy. In practice, this gives rise to a trade-off: to achieve a good approximation, one needs to have a large number of neurons, which makes the implementation of the training algorithm harder.

In this regard, nearly a century of research in approximation theory suggests that the higher the order of smoothness of the target function, the smaller the number of neurons should be, needed to achieve the desired accuracy. There are many different definitions of smoothness that give rise to different estimates. For example, under the condition that the Fourier transform of the target function $f$ satisfies $\disp\int_{\RR^D} |\bsw\hat{f}(\bsw)|d\bsw<\infty$, Barron  \cite{barron1993} proved the existence of a neural network with $\O(\e^{-2})$ neurons that gives  an $L^2([0,1]^D)$ error of $\O(\e)$. While it is interesting to note that this number of neurons is essentially independent of the dimension $D$, the constants involved in the $\O$ term as well as the number of derivatives needed to ensure the condition on the target function may increase with $D$. Several authors have subsequently improved upon such results under various conditions on the activation function as well as the target function so as to ensure that the constants depend polynomially on $D$ (e.g., \cite{kurkova1, kurkova2, tractable} and references therein).

The most commonly understood definition of smoothness is just the number of derivatives of the target function. It is well--known from the theory of $n$-widths that if $r\ge 1$ is an integer, and the only a priori information assumed on the unknown target function  is that it is $r$--times continuously differentiable function, then a stable and uniform approximation to within $\e$ by neural networks must have at least a constant multiple of $\e^{-D/r}$ neurons.  
In \cite{optneur}, we gave an explicit construction for a neural network  that achieves the accuracy of $\e$ using $\O(\e^{-D/r})$ neurons arranged in a single hidden layer. It follows that this suffers from a curse of dimensionality, in that the number of neurons increases exponentially with the input dimension $D$. Clearly, if the smoothness $r$ of the function increases linearly with $D$, as it has to in order to satisfy the condition in \cite{barron1993}, then this bound is also ``dimension independent''.

While this is definitely unavoidable for neural networks with one hidden layer, the most natural way out is to achieve local approximation; i.e., given an input $\x$, construct a network with a uniformly bounded number of neurons that approximates the target function with the optimal rate of approximation near the point $\x$, preferably using the values of the function also in a neighborhood of $\x$. Unfortunately, this can never be achieved as we proved in 
\cite{chui1994neural}. Furthermore,  we have proved in \cite{ chui1996limitations} that even if we allow each neuron to evaluate its own activation function, this local approximation fails. Therefore the only way out is to use a neural network with more than one hidden layer, called \emph {deep net} (for deep neural network). Indeed, local approximation can be achieved by a deep net as proved in our papers \cite{multilayer, mhaskar1993neural}. In this regard, it is of interest to point out that  an adaptive version of \cite{multilayer, mhaskar1993neural} was derived in \cite{lermont} for prediction of time series, yielding as much as 150\% improvement upon the state--of--the--art at that time, in the study of the floor market problem. 

Of course, the curse of dimensionality is inherent to the problem itself, whether with one or more hidden layers. Thus, while it is possible to construct a deep net to approximate a function at each point arbitrarily closely by using a uniformly bounded number of  neurons, the uniform approximation on an entire compact set, such as a cube, would still require an approximation at a number of points in the cube, and this number increasing exponentially with the input dimension. Equivalently, the effective number of neurons for approximation on the entire cube is still exponentially increasing with the input dimension.

In addition to the high dimensionality, another difficulty in solving the function approximation problem is that the data may be not just high dimensional but unstructured and sparse. A relatively recent idea which has been found very useful in  applications, in fact, too many to    list exhaustively, is to consider the points $\x$ as being sampled from an unknown, low dimensional sub--manifold $\XX$ of the ambient high dimensional space $\RR^D$. The understanding of the geometry of $\XX$ is the subject of the bulk of modern research in the area of diffusion geometry. An introduction to this subject can be found in the special issue \cite{achaspissue} of Applied and Computational Harmonic Analysis. 
The basic idea is to construct the so--called diffusion matrix from the data, and use its eigen--decomposition for finding local coordinate charts and other useful aspects of the manifold. The convergence of the eigen--decomposition of the matrices to that of the Laplace--Beltrami and other differential operators on the manifold is discussed, for example, in \cite{belkinfound, lafon, singer}. It is shown in \cite{jones2008parameter, jones2010universal}  that some of the eigenfunctions on the manifolds yield a local coordinate chart on the manifold. In the context of deep learning, this idea is explored as a function approximation problem in \cite{cohen_diffusion_net2015}, where a deep net is developed in order to learn the coordinate system given by the eigenfunctions.  

On the other hand, while much of the research in this direction is focused on understanding the data geometry,  the theoretical foundations for the problems of function approximation and harmonic analysis on such data--defined manifold are developed extensively in \cite{mauropap, frankbern, modlpmz, eignet, heatkernframe, chuiinterp}. The theory is developed more recently for kernel construction on directed graphs and analysis of functions on changing data in our paper \cite{tauberian}. However, a drawback of the approach based on data--defined manifolds, known as the out--of--sample extension problem, is that since the diffusion matrix is constructed entirely using the available data, the whole process must be done again if new data become available. A popular idea is then to extend the eigen--functions to the ambient space  by using the so called Nystr\"om extension \cite{coifman2006geometric}. 

The objective of this present paper is to describe a deep learning approach to the problem of function approximation, using three groups of networks in the deep net. The lowest layer accomplishes dimensionality reduction by learning the local coordinate charts on the unknown manifold \textbf{without using any eigen--decomposition}. Having found the local coordinate system, the problem is reduced to the classical problem of approximating a function on a cube in a relatively low dimensional Euclidean space. For the next two layers, we may now apply the powerful techniques from approximation theory to approximate the target function $f$, given the samples on the training data set $\C$. We describe two approaches to construct the basis functions using multi--layered neural networks, and to construct other networks to use these basis functions in the next layer to accomplish the desired function approximation. 

We summarize some of the highlights of our paper. 
\begin{itemize}
\item We give a very simple learning method for learning the local coordinate chart near each point. The subsequent approximation process is then entirely local to each coordinate patch.
\item Our method allows us to solve the pre--image problem easily; i.e., to generate a point on the manifold  corresponding to a given local coordinate description.
\item The learning method itself \textbf{does not involve} any optimization based technique, except probably for reducing the noise in the values of the function itself.
\item We provide optimal error bounds on approximation based on the smoothness of the function, while the method itself \emph{does not require an a priori knowledge of such smoothness.}
\item Our methods can solve easily the 
out--of--sample extension problem. Unlike the Nystr\"om extension process, our method does not require any elaborate construction of kernels defined on the ambient space and commuting with certain differential operators on the unknown manifold.
\item Our method is designed to control the growth of the out--of--sample extension in a tubular neighborhood of $\XX$, and is local to each coordinate patch.
\end{itemize}

 This paper is organized as follows. In Section~\ref{mainsect}, we describe the main ideas in our approach. The local coordinate system is described in detail in Section~\ref{loccordsect}. Having thus found a local coordinate chart around the input, the problem of function approximation reduces to the classical one. In Section~\ref{locbasissect}, we demonstrate how the popular basis functions used in this theory can be implemented using neural networks with one or more hidden layers. The function approximation methods which work with unstructured data without using optimization are described in Section~\ref{approxsect}. In Section~\ref{extsect}, we explain how our method can be used to solve both the pre--image problem and the out--of--sample extension problem.

\bhag{Main ideas and results}\label{mainsect}
The purpose of this paper is to develop a deep learning algorithm to learn a function $f:\XX\to\RR$, where $\XX$ is a 
$d$ dimensional compact Riemannian sub--manifold  of a Euclidean space $\RR^D$, with $d\ll D$, given \emph{training data} of the form $\{(\x_j, f(\x_j))\}_{j=1}^M$, $\x_j\in\XX$. It is important to note that $\XX$ itself is not known in advance; the points $\x_j$ are known only as $D$--dimensional vectors, presumed to lie on $\XX$. In Sub--section~\ref{ideasect}, we explain our main idea briefly. In Sub--section~\ref{loccordsect}, we derive a simple construction of the local coordinate chart for $\XX$. In Sub--section~\ref{locbasissect}, we describe the construction of a neural network with one or more hidden layers to implement two of the basis functions used commonly in function approximation.  While the well known classical approximation algorithms require a specific placement of the training data, one has no control on the location of the data in the current problem. In Section~\ref{approxsect}, we give algorithms suitable for the purpose of solving this problem.

\subsection{Outline of the main idea}\label{ideasect}

Our approach  is the following.
\begin{enumerate}
\item \label{loccoord} $\XX$ is a finite union of local coordinate neighborhoods, and $\x$ belongs to one of them, say $\mathbb{U}$. We find a local coordinate system for this neighborhood in terms of Euclidean distances on $\RR^D$, say $\Phi : \mathbb{U}\to [-1,1]^d$, where $d$ is the dimension of the manifold. Let $\y=\Phi(\x)$, and with a relabeling for notational convenience, $\{\x_j\}_{j=1}^K$ be the points in $\mathbb{U}$, $\y_j=\Phi(\x_j)$. This way, we have reduced the problem to approximating $g=f\circ\Phi :[-1,1]^d\to\RR$ at $\y$, given the values
$\{(\y_j,g(\y_j))\}_{j=1}^K$, where $g(\y_j)=f(\x_j)$. We note that $\{\y_j\}$ is a subset of the unit cube of low dimensional Euclidean space, representing a local coordinate patch on $\XX$. Thus, the problem of approximation of $f$ on this patch is reduced that of  approximation of $g$, a well studied classical approximation problem.
\item \label{spline} We will summarize the solution to this problem using neural  networks with one or more hidden layers, e.g., an implementation of multivariate tensor product spline approximation using multi--layerd neural network. 
\end{enumerate}
Thus, the layers of our deep learning networks will have three main layers.
\begin{enumerate}
\item The bottom layer receives the input $\x$, figures out which of the points $\x_j$ are in the coordinate neighborhood of $\x$, and computes the local coordinates $\y$, $\y_j$.
\item The next several layers compute the local basis functions necessary for the approximation, for example, the $B$--splines and their translates using the multi--layered neural network as in \cite{multilayer}.
\item The last layer receives the data $\{(\y_j,g(\y_j))\}_{j=1}^K$, and computes the approximation described in Step~\ref{spline} above.
\end{enumerate}

\subsection{Local coordinate learning}\label{loccordsect}
We assume that $1\le d\le D$ are integers, $\XX$ is a 
$d$ dimensional smooth, compact, connected, Riemannian sub--manifold  of a Euclidean space $\RR^D$, with geodesic distance $\rho$. 

Before we discuss our own construction of a local coordinate chart on $\XX$, we wish to motivate the work by describing a result from \cite{jones2010universal}. Let $\{\lambda_k^2\}_{k=0}^\infty$ be the sequence of eigenvalues of the (negative of  the) Laplace--Beltrami operator on $\XX$, and for each $k\ge 0$, $\phi_k$ be the eigenfunction corresponding to the eigenvalue $\lambda_k^2$.
We define a formal ``heat kernel'' by
\be\label{heatkerndef}
K_t(\x,\y)=\sum_{k=0}^\infty \exp(-\lambda_k^2t)\phi_k(\x)\phi_k(\y).
\ee 
The following result is a paraphrasing of the heat triangulation theorem proved in \cite[Theorem~2.2.7]{jones2010universal} under weaker assumptions on $\XX$.

\begin{theorem}\label{jmstheo}{\rm (cf. \cite[Theorem~2.2.7]{jones2010universal})}
 Let $\x_0^*\in\XX$. There exist constants  $R>0$, $c_1,\cdots, c_6>0$ depending on $\x_0^*$ with the following property. Let $\mathbf{p}_1,\cdots,\mathbf{p}_d$ be $d$ linearly independent vectors in $\RR^d$, and $\x_j^*\in\XX$ be chosen so that $\x_j^*-\x_0^*$ is in the direction of $\mathbf{p}_j$, $j=1,\cdots,d$, and 
$$
c_1R\le \rho(\x_j^*,\x_0^*)\le c_2R, \qquad j=1,\cdots,d,
$$ 
and $t=c_3R^2$. Let $B\subset\XX$ be the geodesic ball of radius $c_4R$, centered at $\x_0^*$, and 
\be\label{jmsphidef}
\Phi_{\mbox{jms}}(\x)=R^d(K_t(\x,\x_1^*), \cdots, K_t(\x,\x_d^*)), \qquad \x\in B.
\ee
Then
\be\label{jmsdistpreserve}
\frac{c_5}{R}\rho(\x_1,\x_2)\le \|\Phi_{\mbox{jms}}(\x_1)-\Phi_{\mbox{jms}}(\x_2)\|_d\le 
\frac{c_6}{R}\rho(\x_1,\x_2), \qquad \x_1,\x_2\in B.
\ee
\end{theorem}
Since the paper \cite{jones2010universal} deals with a very general manifold, the mapping $\Phi_{\mbox{jms}}$ is not claimed to be a diffeomorphism, although it is obviously one--one on $B$.

We note that even in the simple case of a Euclidean sphere, an explicit expression for the heat kernel is not known. In practice, the heat kernel has to be approximated using appropriate Gaussian networks \cite{lafon}. In this section, we aim to obtain a local coordinate chart that is computed directly in terms of Euclidean distances on $\RR^D$, and depends upon $d+2$ trainable 
parameters. The construction of this chart constitutes the first hidden layer of our deep learning process. As explained in the introduction, once this chart is in place, the question of function extension on the manifold reduces locally to the well studied problem of function extension on a $d$ dimensional unit cube. 

To describe our constructions,we first develop some notation.

In this section, it is convenient to use the notation $\x=(x^1,\cdots,x^D)\in\RR^D$ rather than $\x=(x_1,\cdots,x_D)$, which we will use in the rest of the sections. If $1\le d\le D$ is an integer, and $\x\in\RR^d$, $\|\x\|_d$ denotes the Euclidean norm of $\x$. If $\x\in\RR^D$, we will write $\pi_c(\x)=(x^1,\cdots,x^d)$, $\|\x\|_d=\|\pi_c(\x)\|_d$. If $\x\in\RR^d$, $r>0$,
$$
B(\x,r)=\{\y\in\RR^d : \|\x-\y\|_d\le r\}.
$$

There exists $\delta^*>0$ with the following properties. The manifold is covered by finitely many geodesic balls such that for the center $\x_0^*\in\XX$ of any of these balls, there exists a diffeomorphism, namely, the exponential coordinate map $u=(u^1,\cdots,u^D)$ from $B_d(0,\delta^*)$ to the geodesic ball around $\x_0^*=u(0)$ \cite[p.~65]{docarmo_riemannian}. If $J$ is the Jacobian matrix for $u$, given by $J_{i,j}(\y)=D_iu^j(\y)$, $\y\in B_d(0,\delta^*)$, then 
\be\label{finaljacobiatzero}
J(0)=[I_d|0_{d,D-d}].
\ee
Further, there exists $\kappa>0$ (independent of $\x^*$) such that
\be\label{jacobimodcont}
\|J(\mathbf{q})-J(0)\|\le \kappa\|\mathbf{q}\|_d, \qquad \mathbf{q}\in B_d(0,\delta^*).
\ee
Let $\eta^*:=\min(\delta^*, 1/(2\kappa))$. Then \eref{jacobimodcont} implies that 
\be\label{jacobinormest}
1/2\le 1-\kappa\|\mathbf{q}\|_d\le \|J(\mathbf{q})\|\le 1+\kappa\|\mathbf{q}\|_d\le 2, \qquad \mathbf{q}\in B_d(0,\eta^*).
\ee
In turn, this leads to
\be\label{rhoeuccomp1}
(1/2)\rho(u(\mathbf{p}),u(\mathbf{q}))\le \|\mathbf{p}-\mathbf{q}\|_d\le 2\rho(u(\mathbf{p}), u(\mathbf{q})), \qquad \mathbf{p},\mathbf{q}\in B_d(0,\eta^*).
\ee

Let $\x_\ell^*=u(\mathbf{q}_\ell)$, $\ell=1,\cdots,d$, be chosen with the following properties:
\be\label{nbdcond}
\|\mathbf{q}_\ell\|_d\le \eta^*, \qquad \ell=1,\cdots,d,
\ee
and, with the matrix function $U$ defined by 
\be\label{umatrixdef}
U_{i,j}(\mathbf{q})=u^i(\mathbf{q})-(\x_j^*)^i,
\ee
 we have
\be\label{indepcond}
\|J(0)U(0)\y\|_d\ge \gamma>0, \qquad \|\y\|_d= 1.
\ee
Any set $\{\x_\ell^*\}$ with these properties will be called  \emph{coordinate stars} around $\x^*$. We note that the matrix $J(0)U(0)$ has columns given by $\pi_c(\x^*-\x_j^*)$, $j=1,\cdots,d$, and hence, can be computed without reference to the map $u$. Let 
\be\label{betastardef}
\beta^* := (1/2)\min\left(\frac{1}{2\kappa}, \delta^*,\frac{\gamma}{8\sqrt{d}}\right).
\ee 

\begin{theorem}\label{loccordtheo}
Let $\Psi(\mathbf{q}):=(\|u(\mathbf{q})-u(\mathbf{q}_\ell)\|^2_D)_{\ell=1}^d\in\RR^d$. Then \\
{\rm (a)} $\Psi$ is a diffeomorphism on $B_d(0,2\beta^*)$. If $\mathbf{p}, \mathbf{q}\in B_d(0,2\beta^*)$, $\x=u(\mathbf{p})$, $\y=u(\mathbf{q})$, then
\be\label{distortest}
\frac{\gamma}{2}\rho(\x,\y)\le \|\Psi(\mathbf{p})-\Psi(\mathbf{q})\|_d \le 32\sqrt{d}\eta^*\rho(\x,\y).
\ee
{\rm (b)} The function $\Psi$ is a diffeomorphism from $B_d(0,\beta^*)$ onto $B_d(\Psi(0),\beta^*)$. 
 \end{theorem}
 
\begin{rem}\label{maptocuberem}
{\rm
Let $\mathbb{B}=u(B_d(0,\beta^*))\subset \XX$ be a geodesic ball around $\x_0^*$. For $\x\in \mathbb{B}$, we define
$$
\phi(\x)=\Psi(u^{-1}(\x))=(\|\x-\x_\ell^*\|_D^2).
$$
Then Theorem~\ref{loccordtheo}(b) shows that 
$\phi$ is a diffeomorphism from $\mathbb{B}$ onto $B_d(\Psi(0),\beta^*)$. Since $\Psi(0)=(\|\x_0^*-\x_\ell^*\|_D^2)$, 
$$
\Phi(\x)=\frac{\sqrt{d}}{\beta^*}(\phi(\x)-\Psi(0)), \qquad \x\in \mathbb{B}
$$
maps $\mathbb{B}$ diffeomorphically onto $B_d(0,\sqrt{d})\supset [-1,1]^d$. Let $\mathbb{U}=\Phi^{-1}([-1,1]^d)$. Then  $\mathbb{U}$ is a neighborhood of $\x_0^*$ and $\Phi$ maps $\mathbb{U}$ diffeomorphically onto $[-1,1]^d$. We oberve that $\XX$ is a union of finitely many neighborhoods of the form $\mathbb{U}$, so that any $\x\in\XX$ belongs to at least one such neighborhood. Moreover, $\Phi(\x)$ can be computed entirely in terms of the description of $\x$ in terms of its $D$--dimensional coordinates. \qed
}
\end{rem}

\begin{rem}\label{trainrem}
{\rm The trainable parameters are thus $\beta^*$, and the points $\x_0^*, \cdots, \x_d^*$. Since $\|J(0)\|=1$, the condition \eref{indepcond} is satisfied if $\x_\ell^*-\x_0^*$ are along linearly independent directions as in Theorem~\ref{jmstheo}.
\qed}
\end{rem}

\begin{rem}\label{networkimprem}
{\rm
Since the mapping $\Phi$ in Remark~\ref{maptocuberem} is a quadratic polynomial in $\x$, it can be implemented as a neural network with a single hidden layer using the activation function given in \eref{requdef} as described in Sub--Section~\ref{polysubsect}.
}
\end{rem}

\begin{uda}\label{helixexample}
{\rm
Let $0<a<1$, and $M=M_a$ be the circular helix defined by
$$
\u(s)=(\cos as, \sin as, \sqrt{1-a^2}s)^T.
$$
Clearly, $M$ is a one dimensional manifold, and $s$ is the arclength parameter, measured from $(1,0,0)^T$. The curvature at any point is $a^2$. For any point $\z_0\in M$, $U_{\z_0}=M$, with the diffeomorphism given by $\u(\tan^{-1}(\pi t/2))$, $t\in (-1,1)$. An interesting fact is that $\|\u(s+2\pi)-\u(s)\|_3=2\pi\sqrt{1-a^2}$ can be made arbitrarily small by choosing $a$ close to $1$, even though the geodesic distance between $\u(s+2\pi)$ and $\u(s)$ is $2\pi$. Let $s_0\in\RR$, $s_0+\pi/4\le s_1\le s_0+3\pi/8$, and $U := \{\u(s)\ :\ |s-s_0|\le \pi/8\}$. Let $\Psi(s):=\|\u(s)-\u(s_1)\|_3^2$. It is easy to calculate that
$$
\Psi(s)= 2-2\cos (a(s-s_1))+(1-a^2)(s-s_1)^2,
$$
so that
$$
\Psi'(s)=2a\sin (a(s-s_1))+ 2(1-a^2)(s-s_1).
$$
If $\u(s)\in U$, then $\pi/8\le s_1-s\le \pi/2$. Therefore, using the well known estimates
$$
\frac{2\theta}{\pi} \le \sin (\theta) \le \theta, \qquad \theta\in [0,\pi/2],
$$
we obtain that
\be\label{snakeupbd}
|\Psi'(s)| \le 2a^2(s_1-s)+2(1-a^2)(s_1-s)\le \pi,
\ee
and
\be\label{snakelowbd}
|\Psi'(s)| \ge \frac{4a^2}{\pi}(s_1-s) + 2(1-a^2)(s_1-s)\ge \frac{\pi}{8}(2-2a^2(1-2/\pi))\ge 1/2.
\ee
Hence, for any points $\u(t_1), \u(t_2)\in U$, we have
$$
(1/2)|t_1-t_2|\le \left|\|\u(t_1)-\u(s_1)\|_3^2-\|\u(t_2)-\u(s_1)\|_3^2\right|\le \pi |t_1-t_2|.
$$
We note that  the neighborhood $U$ where this estimate holds and the constants are independent of the curvature.
\qed
}
\end{uda}
The remainder of this section is devoted to the proof of Theorem~\ref{loccordtheo}.

\begin{lemma}\label{umatrixlemma}
 Let $\mathbf{q} \in B_d(0,\eta^*)$.  Each of the following statements hold for the matrix $U$ defined in \eref{umatrixdef}:
\be\label{ucontinuity}
 \|U(\mathbf{q})-U(0)\| \le 2\sqrt{d}\|\mathbf{q}\|_d,
\ee
\be\label{uupest}
\|U(\mathbf{q})\|\le 2\sqrt{d}\max_{1\le\ell\le d}\|\mathbf{q}-\mathbf{q}_\ell\|_d\le 4\sqrt{d}\eta^*,
\ee
\be\label{scriptjcont}
\|J(\mathbf{q})U(\mathbf{q})-J(0)U(0)\| \le 4\sqrt{d}\|\mathbf{q}\|_d.
\ee
With $\beta^*$ as in \eref{betastardef}, for $\|\mathbf{q}\|_d\le 2\beta^*$, $\|\y\|_d=1$,
\be\label{ulowest}
\|J(\mathbf{q})U(\mathbf{q})\y\|_d\ge  \gamma/2.
\ee
\end{lemma}

\begin{Proof}\ 
In view of \eref{jacobinormest} and the mean value theorem, we have for $\|\mathbf{p}\|_d\le \eta^*$,
\be\label{pf1eqn2}
 \|u(\mathbf{q})-u(\mathbf{p})\|_D \le 2\|\mathbf{q}-\mathbf{p}\|_d.
\ee

We observe further that for any integers $m, \ell$, $U(\mathbf{q})_{m,\ell}-U(0)_{m,\ell}= u^m(\mathbf{q})-u^m(0)$. Consequently, for any $\y\in\RR^d$, $\|\y\|_d\le 1$,
$$
\|(U(\mathbf{q})-U(0))\y\|_D=\|u(\mathbf{q})-u(0)\|_D\sum_{\ell=1}^d |y^\ell|\le 2\sqrt{d}\|\mathbf{q}\|_d\|\y\|_d.
$$
This proves \eref{ucontinuity}.

In view of \eref{pf1eqn2}, used with $\mathbf{q}_\ell$ in place of $\mathbf{p}$, $\ell=1,\cdots,d$, we obtain for all $\y\in\RR^d$, $\|\y\|_d\le 1$,
$$
\left\|\sum_{\ell=1}^d \y^\ell (u(\mathbf{q})-u(\mathbf{q}_\ell) )\right\|_D \le \sum_{\ell=1}^d |\y^\ell|\|u(\mathbf{q})-u(\mathbf{q}_\ell)\|_D \le 2\sqrt{d}\max_{1\le \ell\le d}\|\mathbf{q}-\mathbf{q}_\ell\|_d.
$$
This proves \eref{uupest}.

In view of \eref{jacobimodcont}, \eref{uupest}, \eref{ucontinuity}, we obtain for $\|\mathbf{q}\|_d\le \eta^*\le 1/(2\kappa)$ that 
\begin{eqnarray*}
\|J(\mathbf{q})U(\mathbf{q})-J(0)U(0)\|&=&\|(J(\mathbf{q})-J(0))U(\mathbf{q}) + J(0)(U(\mathbf{q})-U(0))\|\\
&\le& \|J(\mathbf{q})-J(0)\|\|U(\mathbf{q})\| +\|J(0)\|\|U(\mathbf{q})-U(0)\| \\
&\le& 4\sqrt{d}\eta^*\kappa\|\mathbf{q}\|_d + 2\sqrt{d}\|\mathbf{q}\|_d= 2\sqrt{d}(1+2\eta^*\kappa)\|\mathbf{q}\|_d\le 4\sqrt{d}\|\mathbf{q}\|_d.
\end{eqnarray*}
This proves \eref{scriptjcont}. The estimate \eref{ulowest} follows easily from this and \eref{indepcond}.
\end{Proof}

\noindent
\textsc{Proof of Theorem~\ref{loccordtheo}.}
In this proof only, let  ${\cal J}(\mathbf{q})$ be the Jacobian of $\Psi$: ${\cal J}_{i,j}(\mathbf{q})=D_i(\|u(\mathbf{q})-u(\mathbf{q}_j)\|^2)$. Then ${\cal J}(\mathbf{q})=2J(\mathbf{q})U(\mathbf{q})$. The estimate \eref{indepcond} shows that ${\cal J}(0)$ is invertible, and that $\|{\cal J}(0)^{-1}\|\le 1/(2\gamma)$. The estimate \eref{scriptjcont} then shows that
\be\label{pf2eqn1}
\|{\cal J}(\mathbf{q})-{\cal J}(0)\| \le 1/(2\|{\cal J}(0)^{-1}\|), \qquad \|\mathbf{q}\|_d\le 2\beta^*.
\ee
Therefore, the inverse function theorem as given in \cite[Theorem~9.24 and its proof]{rudin1976principles} implies that $\Psi$ is a diffeomorphism on $B(0,2\beta^*)$ as claimed. For $\|\mathbf{q}\|_d\le 2\beta^*$, \eref{ulowest} shows that $\|{\cal J}(\mathbf{q})^{-1}\|\le  1/\gamma$. Also, \eref{uupest} and \eref{jacobinormest}  show that $\|{\cal J}(\mathbf{q})\|\le 16\sqrt{d}\eta^*$. Hence, the mean value theorem implies that
\be\label{pf2eqn2}
\gamma \|\mathbf{q}-\mathbf{p}\|_d\le \|\Psi(\mathbf{q})-\Psi(\mathbf{p})\|_d \le 16\sqrt{d}\eta^* \|\mathbf{q}-\mathbf{p}\|_d.
\ee
Together with \eref{rhoeuccomp1}, this implies \eref{distortest}.

The part (b) follows also from \cite[Theorem~9.24 and its proof]{rudin1976principles} and \eref{pf2eqn1}. 
\qed

 \subsection{Local basis functions}\label{locbasissect}
Having found a local coordinate map $\Phi$ on a neighborhood $\mathbb{U}$ of $\x$ on $\XX$, the problem of extending $f$ from $\{\x_j\}\cap \mathbb{U}$ to $\mathbb{U}$ is reduced to extending $f\circ\Phi$ from $\C=\{\y_j=\Phi(\x_j)\}\subset [-1,1]^d$, a classical approximation problem. There is, of course, 100+ years of research on this subject. We restrict ourselves to two examples, which can be implemented using neural networks with one or more hidden layers.  One of the most popular activation function in the deep learning literature 
(e.g., \cite{lecun2015deep}) is the 
rectified linear unit function
$$
t_+=\max(0,t).
$$
Since this function is not continuously differentiable, there are some technical difficulties to use common algorithms like back--propagation with these activation functions. Although we do not need back--propagation in our theory, we prefer to deal with a \emph{rectified quadratic unit function} defined for $t\in\RR$ by
\be\label{requdef}
\sigma(t)=\left\{\begin{array}{ll}
t^2, &\mbox{ if $t\ge 0$,}\\
0, &\mbox{ if $t<0$,}
\end{array}\right.
\ee
which is continuously differentiable on $\RR$. Our theory will work in general with any activation function  of order $k\ge 2$; i.e., with a function $\sigma$ that satisfies
\be\label{actfnorder}
\lim_{t\to -\infty}\frac{\sigma(t)}{t^k}=0, \qquad \lim_{t\to \infty}\frac{\sigma(t)}{t^k}=1,
\ee
but for the sake of clarity of exposition, we will use only the activation function $\sigma$ defined in \eref{requdef}.

\subsection{Polynomials}\label{polysubsect}
The most basic class of classical approximants is the set of all polynomials. For $n>0$, we denote the class of all algebraic polynomials of coordinatewise degree at most $n$ in $d$ variables by $\Pi_n^d$. (It is convenient to use the same notation also when $n$ is not an integer; in this case, $\Pi_n^d$ is just $\Pi_{\lfloor n\rfloor}^d$.

The basic implementation of polynomials is given in \cite[Proof of Theorem~3.1]{chuili1992}, where an explicit construction is given for finding the weights $\{\w_k\}$, the thresholds $\{b_k\}$ and the coefficients $\{a_k\}$ used in \eref{chuilipolynet} below.

\begin{theorem}\label{chuilitheo}
Let $n>0$,  $N=2^{\lceil \log_2 n\rceil}$,
$P\in\Pi_N^d\supseteq \Pi_n^d$,
then there exist weights $\w_1,\cdots,w_{\mbox{dim}(\Pi_N^d)}$ and real numbers $a_1,\cdots, a_{\mbox{dim}(\Pi_N^d)}$, $b_1,\cdots, b_{\mbox{dim}(\Pi_N^d)}$ such that
\be\label{chuilipolynet}
P(\x)=\sum_{k=1}^{\mbox{dim}(\Pi_N^d)}a_k(\w_k\cdot\x +b_k)^N, \qquad x\in \RR^d.
\ee
Here, the weights $\{\w_k\}$ and the thresholds $\{b_k\}$ are independent of $P$ and the coefficients $\{a_k\}$ are linear functionals on $\Pi_N^d$.
\end{theorem}

We observe that
\be\label{powerfromtruncpow}
t^2=\sigma(t)+\sigma(-t), \qquad x\in\RR,
\ee
while
\be\label{iterpower}
t^N=((t^2)^2\cdots)^2, (\log_2 N \mbox{ times}),
\ee
so that 
the expression on the right hand side of \eref{chuilipolynet} can be expressed as a neural network with $\log_2 N$ hidden layers. 

We note that a neural network with one hidden layer  is given in \cite{optneur}, but using a $C^\infty$ activation function $\sigma$; e.g.,
$\sigma(t)= (1+e^{-t})^{-1}$. This uses the fact that for $\w,\x\in\RR^d$ and $b\in\RR$, such that none of the derivatives $\sigma^{(j)}(b)$, $j=0,1,\cdots$, equal to $0$,
\be\label{optneurformula}
\frac{\partial^{|\mathbf{k}|}}{\partial \mathbf{w}^{\mathbf{k}}}\sigma(\mathbf{w}\cdot\x+b)\bigg|_{\mathbf{w}=0} =\x^{\mathbf{k}}\sigma^{|\mathbf{k}|}(b).
\ee
A finite difference scheme to implement this differentiation yields a neural network with one hidden layer, containing exactly $\mbox{dim}(\Pi_n^d)$ neurons, and should be stable for $C^\infty$ functions. If stability is a greater concern, then one may use other  numerical differentiation schemes to implement this formula, e.g., spectral methods \cite{gottlieb1977numerical}. 

\subsection{$B$--splines}\label{splinesubsect}

For $t\in\RR$, and integer $m\ge 1$, let 
$$
t^m_+=\left\{\begin{array}{ll}
t^m, &\mbox{ if $t\ge 0$,}\\
0, &\mbox{otherwise.}
\end{array}\right.
$$
A tensor product cardinal $B$-spline at $\y\in [-1,1]^d$ is defined by
\be\label{tensbsplinedef}
N_m(\y)=\frac{1}{(m-1)!^d}\sum_{\k\in\ZZ, \atop \k\ge 0,\ |\k|_\infty\le m} (-1)^{|\k|_1}\prod_{j=1}^d {m\choose k_j}\prod_{j=1}^d (y_j-k_j)^m_+.
\ee

It is explained in \cite{multilayer,mhaskar1993neural} that the quantity $N_m(\y)$ can be computed using a neural network with a sigmoidal function of order $m-1$ consisting of finitely many neurons arranged in multiple hidden layers (the number of neurons and layers depending on $m$ and $d$ alone). 
Thus, if $m$ is a power of $2$, then each of the terms $(y_j-k_j)^m_+$ can be implemented as an iterated power of $(y_j-k_j)^2_+$ (cf. \eref{iterpower}.) The product of $d$ such expressions can be implemented using either Theorem~\ref{chuilitheo} as a network with mulitple hidden layer and utilizing the rectified quadratic unit function as the activation function, or a discretization of the formula \eref{optneurformula} using a $C^\infty$ sigmoidal function as explained in Sub--section~\ref{polysubsect}.

\subsection{Function approximation}\label{approxsect}
In this section, if $\y\in\RR^d$, $\|\y\|_\infty$ is the $\ell^\infty$ norm of $\y$.
\subsubsection{Spline based approximation}\label{splineapproxsect}

In \cite[Section~4.5]{chuiwaveletbk},  \cite{chui_diamond_quasi_int90}, a quasi--interpolatory spline function is defined by
\be\label{quasisplinedef}
Q_m(f)(\y)=\sum_{\k\in\ZZ^d}\lambda_m^*(f(\cdot+\k))N_m(\y+m/2+\k),
\qquad \y\in\RR^d,
\ee
where $\lambda_m^*$ are compactly supported linear functionals, designed specifically to ensure that $Q_m(P)=P$ for every polynomial $P$ of coordinatewise degree at most $m-1$ in $d$ variables. With $Q_{m,h}(f)(\y)=Q_m(f(h(\cdot)))(\y/h)$, $h>0$, one has the approximation bound for small $h$:
\be\label{quasi_spline_approx_bd}
\max_{\y\in [-1,1]^d}|Q_{m,h}(f)(\y)-f(\y)|=\O(h^m).
\ee
The linear functionals $\lambda_m^*$ are based on \emph{finitely many} samples of $f$ at the grid points in a compact subset of $\ZZ^d$. In our context, the data for approximating $f$ is not in this form. Therefore, we may use the following algorithm given in \cite{quasiint2000}, where we assume that $\lambda_m^*$ is scaled so as to be supported on $[-1,1]^d$.
\begin{quotation}
\noindent {\bf Given:} A  set $\C=\{\bs{\xi}_j\}$ of points in $[-1,1]^d$. Let
$$
\delta(\C)=\max_{\bs{\xi}_1-\bs{\xi}_2\in\C}|\bs{\xi}_1-\bs{\xi}_2|_\infty,
$$
and $\delta(\C)$ be sufficiently small.\\

\noindent {\bf Objective:} To find real numbers
$\{a_{\bs{\xi}}\}_{\bs{\xi}\in\C}$ such that the functional
\be\label{splinequad}
\gamma(f) := \sum_{\bs{\xi}\in\C} a_{\bs{\xi}} f(\bs{\xi})
\ee
satisfies
$$
\gamma(P) = \lambda_m^*(P), \qquad \mbox{ if $P$ is $d$--variate polynomial of coordinatewise degree $\le m-1$.}
$$
{\bf Steps:}
\begin{enumerate}
\item Divide $[-1,1]^d$ into congruent subcubes of side not exceeding $2\delta(\C)$. 
\item Choose  $\C_0\subseteq \C$, so that each subcube has exactly one point of $\C_0$.
\item Solve the following (underdetermined) system of equations for
the unknowns $a_{\bs{\xi}}$, $\bs{\xi}\in\C_0$.

\be\label{axieqn}
\sum_{\bs{\xi}\in\C}a_{\bs{\xi}} \bs{\xi}^\k = \lambda_m^*((\cdot)^\k), \qquad |\k|_\infty\le m-1.
\ee

\item Set $a_{\bs{\xi}}:=0$ if $\bs{\xi}\in\C\setminus \C_0$.
\item Output $(a_{\bs{\xi}})_{\bs{\xi}\in\C}$.
\end{enumerate}
\end{quotation}
Substituting $\gamma$ in place of $\lambda_m^*$ in the definition of $Q_m(f)$ yields the desired spline approximation
\be\label{scat_quasi_spline_def}
\tilde{Q}_m(f)(\y)=\sum_{\k\in\ZZ^d}\gamma(f(\cdot+\k))N_m(\y+m/2+\k),
\qquad \y\in\RR^d,
\ee
and it is proved in \cite{quasiint2000} that the estimate \eref{quasi_spline_approx_bd} holds with $\tilde{Q}_m$ replacing $Q_m$. 

\subsubsection{Polynomial quasi--interpolation}\label{polyapproxsect}

A standard method for polynomial approximation is to consider a filtered projection defined in \eref{summopdef} below. 

The Chebyshev polynomials (of first kind) are defined recursively for $t\in\RR$ and integer $m\ge 0$ by
\be\label{chebpolydef}
T_0(t)=1,\quad T_1(t)=t,\quad T_m(t)=2tT_{m-1}(t)-T_{m-2}(t).
\ee
In terms of monomials, the Chebyshev polynomials are given by
\be\label{chebmonoexp}
T_{2n}(t)=\sum_{j=0}^n \frac{(-4)^j}{(2j)!}\prod_{\ell=1}^j (n^2-(j-\ell)^2))t^{2j},\quad T_{2n+1}(t)=\sum_{j=0}^n \frac{(-1)^j(2n+1)^2}{(2j+1)!}\prod_{\ell=1}^j ((2n+1)^2-(2j-2\ell+1)^2)t^{2j+1}.
\ee
For $\y\in\RR^d$ and multi--integer $\mathbf{m}=(m^1,\cdots,m^d)\ge 0$, the tensor product Chebyshev polynomial is defined by
\be\label{tenschebpolydef}
T_{\mathbf{m}}(\y)=\prod_{j=1}^d T_{m^j}(y^j).
\ee

We choose a smooth low pass filter $h$; i.e., an even function $h :\RR\to [0,1]$ such that $h(u)=1$ if $|u|\le 1/2$, and $h(u)=0$ if $|u|\ge 1$, and abuse the notation as usual to define
$$
h(\u)=\prod_{j=1}^d h(u^j).
$$
With this filter, we define the kernel
\be\label{kerndef}
\Phi_n(\y,\t)=\sum_{\k\in \ZZ^d} h(\k/n)T_\k(\y)T_\k(\t), \qquad n>0, \ \y,\t\in [-1,1]^d.
\ee
Then the filtered projection operator is defined by
\be\label{summopdef}
V_n(f)(\y)=\int_{[-1,1]^d} f(\t)\Phi_n(\y, \t)\frac{d\t}{\sqrt{(1-(t^1)^2)\cdots (1-(t^d)^2)}}, \qquad \y\in [-1,1]^d,
\ee
It is well known that if $f$ is any continuous function on $[-1,1]^d$, then $\{V_n(f)\}$ converges uniformly to $f$ at the near optimal rate of approximation. For example, if $f$ has partial derivatives up to order $r$ in each variable, then analogously to \eref{quasi_spline_approx_bd}, but for large $n$ rather than small $h$,
\be\label{goodapproxbd}
\max_{\y\in [-1,1]^d}|V_n(f)(\y)-f(\y)| =\O(n^{-r}).
\ee
Theoretically, the question then is to compute $V_n(f)$ using the data $\C$ as in Sub--section~\ref{splineapproxsect}. The procedure we describe below from \cite{indiapap, jaenpap, mnpdiabetes} also describes the choice of the parameter $n$ depending upon the data. 
\begin{quotation}
\noindent {\bf Given:} A  set $\C=\{\bs{\xi}_j\}$ of points in $[-1,1]^d$. Let
$$
\delta^\circ(\C)=\max_{\bs{\xi}_1,\bs{\xi}_2\in\C}\max_{1\le j\le d}|\arccos(\xi^j_1)-\arccos(\x^j_2)|.
$$
and $\delta^\circ(\C)$ be sufficiently small. We also fix an integer $n>0$.\\

\noindent {\bf Objective:} To find real numbers
$\{w_{\bs{\xi}}\}_{\bs{\xi}\in\C}$ such that the functional
\be\label{polyquad}
\gamma^\circ(f) := \sum_{\bs{\xi}\in\C} w_{\bs{\xi}} f(\bs{\xi})
\ee
satisfies
$$
\gamma^\circ(P) = \int_{[-1,1]^d}\frac{P(\t)d\t}{\sqrt{(1-(t^1)^2)\cdots (1-(t^d)^2)}}, 
$$
for all  $d$--variate polynomials $P$ of coordinatewise degree $\le n-1$.\\

\noindent{\bf Steps:}
\begin{enumerate}
\item Divide $[-1,1]^d$ into congruent subcubes of side not exceeding $2\delta^\circ(\C)$. 
\item Choose  $\C_0\subseteq \C$, so that each subcube has exactly one point of $\C_0$.
\item Solve the following (underdetermined) system of equations for
the unknowns $w_{\bs{\xi}}$, $\bs{\xi}\in\C_0$.

\be\label{wxieqn}
\sum_{\bs{\xi}\in\C}w_{\bs{\xi}} T_\k(\bs{\xi}) = \int_{[-1,1]^d}\frac{T_\k(\t)d\t}{\sqrt{(1-(t^1)^2)\cdots (1-(t^d)^2)}}=\delta_{\k,0}, \qquad |\k|_\infty\le n-1.
\ee

\item Set $w_{\bs{\xi}}:=0$ if $\bs{\xi}\in\C\setminus \C_0$.
\item Output $(w_{\bs{\xi}})_{\bs{\xi}\in\C}$.
\end{enumerate}
\end{quotation}
It is proved in the papers cited above that with the discretized operator
\be\label{discsummopdef}
V_{n,\C}(f)(\y)=\sum_{\bs{\xi}\in\C}w_{\bs{\xi}}f(\bs{\xi})\Phi_n(\y,\bs{\xi}),
\ee
one obtains the near best rates of approximation. In particular, if $f$ has partial derivatives up to order $r$ in each variable, then \eref{goodapproxbd} holds with $V_{n,\C}(f)$ replacing $V_n(f)$. In practice, we choose $n$ to be the largest integer such that either the condition number of the system of equations in \eref{wxieqn} is ``reasonable'' or else by checking the resulting errors in \eref{wxieqn} \cite{quadconst}.

The formula \eref{discsummopdef} can be re--written in the form
\be\label{disc_summ_four}
V_{n,\C}(f)(\y)=\sum_{|\k|_\infty\le n-1} h(\k/n)\hat{f}(\C,\k)T_\k(\y),
\ee
where
\be\label{disc_four_coeff}
\hat{f}(\C,\k)=\sum_{\bs{\xi}\in \C} w_{\bs{\xi}}f(\bs{\xi})T_\k(\bs{\xi}).
\ee
Therefore, rather than evaluating the Chebyshev polynomials as defined in \eref{tenschebpolydef}, \eref{chebmonoexp}, one can use a 
multi--layered network to evaluate $V_{n,\C}(f)$ in a more stable manner as follows. The first layer computes the coefficients $h(\k/n)\hat{f}(\C,\k)$ using the available data. The output of this layer is input to a recurrent network to execute a multi--variate version of the well known  Clenshaw algorithm \cite[pp.~78--80]{gautschibk}.

\bhag{Extensions}\label{extsect}

Since the starting point of diffusion geometry is to consider eigen--decomposition of a diffusion matrix, which is constructed using the available data, the entire computation needs to be redone if a new data becomes available. Since the manifold $\XX$ is only an abstract model, it is not even clear that the new data will belong to this manifold. This gives rise to two related questions. One is to find new points on the manifold; the so called pre--image problem \cite{cohen_diffusion_net2015} and the other is the out-of-sample extension problem; i.e., extend the target function to points not necessarily on $\XX$. In this section, we make some comments on how to use the theory described in the previous sections for solving these problems. 

\subsection{Pre--image problem}\label{preimagesect}

 In Remark~\ref{maptocuberem}, we have given an onto diffeomorphism $\Phi : \mathbb{U} \to [-1,1]^d$ where  $\mathbb{U}\subset \XX$.  The pre--image problem is the following. Given a point $\y\in [-1,1]^d$, find $\x\in\XX$ such that $\Phi(\x)=\y$.  This amounts to approximating the $D$--output function $\Phi^{-1}$ on $[-1,1]^d$, given its values at the known points $\{\Phi(\bs{\xi}) : \bs{\xi}\in \C\}$, and can therefore be solved using any of the techniques described in the previous sections.
 
\subsection{Out of sample extension}\label{oosesect}

One well known strategy for function extension outside the manifold is the following. One starts with a compact, positive--semi--definite symmetric kernel $K :\RR^D\times \RR^D$ and considers the eigen--decomposition of $K$ restricted to $\XX\times \XX$; thus, for example, if $\mu^*$ is the volume measure on $\XX$, one finds numbers $\lambda_k\ge 0$ and orthonormal functions $\phi_k$ on $\XX$ such that
\be\label{kerneleigdef}
\int_\XX K(\x,\y)\phi_k(\y)d\mu^*(\y)=\lambda_k\phi_k(\x), \qquad k=0,1,\cdots, \ \x\in\XX.
\ee
A function on $\XX$ can then be expanded in terms of the orthonormal system of functions $\{\phi_k\}$. The extension to $\RR^D$ is achieved by treating \eref{kerneleigdef} as a definition of $\phi_k$ on $\RR^D$ (which can be done since $K$ is defined on $\RR^D\times \RR^D$), and the expansion of the original function $f$ where the basis functions are now interpreted as extended by \eref{kerneleigdef} as the desired extension. This leads to a variety of theoretical problems related to the judicious construction of kernels defined on the whole space whose eigenfunctions are meaningful as functions on $\XX$ (e.g., kernels that commute with the Laplace--Beltrami operator) so as to allow such a construction. In the end, it is not clear how well this extension will behave outside of $\XX$.

Our construction gives an alternative method for extending a function on $\XX$ to a tubular neighborhood of $\XX$, which 
we feel is more appropriate for most applications rather than trying to extend the function to the entire ambient space. Toward this goal, we first explain the local coordinate learning phase for tubular neighborhood of $\XX$. 

Let $s\ge d$ be an integer, $s\le D$. For $\mathbf{q}\in\RR^s$ (or $\mathbf{q}\in\RR^D$), we write $\pi_c(\mathbf{q})=(q^1,\cdots,q^d)$, and
$$
v(\mathbf{q})=u(\pi_c(\mathbf{q}))+(\underbrace{0,\cdots,0}_{d \mbox{ times}},q^{d+1},\cdots,q^s,\underbrace{0,\cdots,0}_{D-s \mbox{ times}})\in\RR^D.
$$
If $J_s$ is the Jacobian matrix for $v$, i.e., $(J_s)_{i,j}(\mathbf{q})=D_iv^j(\mathbf{q})$, then it is easy to check that
\be\label{extjacobi}
J_s(0)=[I_s|0_{s,D-s}],
\ee
and
\be\label{extjacobidiff}
(J_s(\mathbf{q})-J_s(\mathbf{p}))(z)=(J_d(\pi_c(\mathbf{q}))-J_d(\pi_c(\mathbf{p})))(z), \qquad \mathbf{p},\mathbf{q}\in B_d(0,\delta^*), \ z\in\RR^D.
\ee
Consequently, 
\be\label{extkappa}
\|J_s(\mathbf{q})-J_s(0)\|\le \kappa\|\pi_c(\mathbf{q})\|_d\le \kappa\|\mathbf{q}\|_s.
\ee
If we now define for $\mathbf{p}, \mathbf{q}\in B_d(0,\delta^*)$
$$
\rho_1(v(\mathbf{p}),v(\mathbf{q}))^2:=\rho(u(\pi_c(\mathbf{p})),u(\pi_c(\mathbf{q})))^2 +\sum_{k=d+1}^s |p^k-q^k|^2,
$$
then following the same argument as the one leading to \eref{rhoeuccomp1} leads to
$$
(1/2)\rho_1(v(\mathbf{p}),v(\mathbf{q}))\le \|\mathbf{p}-\mathbf{q}\|_s\le 2\rho_1(v(\mathbf{p}), v(\mathbf{q})), \qquad \|\mathbf{p}\|_s, \|\mathbf{q}\|_s\le \eta^*.
$$

Thus, there is no loss of generality in assuming that $\XX$ is already a $s$ dimensional submanifold of $\RR^D$, defined by $v$, and with geodesic distance $\rho_1$. This has several consequences. Even if one overestimates the dimension of the original manifold to be $s$ rather than $d$, the resulting ``distance respecting'' coordinate system will also be ``distance respecting'' for the original manifold, except for the presumably small error resulting from the overestimate. If we have no information about $d$ (or $s$), we may take $s=D$. This would answer the question regarding points off the manifold, as well as take noise into account. However, then the advantage of dimension reduction is lost. Also, all the constants will depend upon $D$ rather than $s$ (or $d$). 

Having defined a local coordinate system for the tubular neighborhood of $\XX$ in this way, we can then construct the local basis functions on this neighborhood as in Section~\ref{locbasissect}.  However, since the original data $\C$ is not dense on the local coordinate patch in the tubular neighborhood, one cannot use the ideas in Section~\ref{approxsect}. We would like instead to keep some control on the growth of the extension operator. For this purpose, we propose to use the minimal Sobolev norm (MSN) interpolant introduced in \cite{bdint} and used very fruitfully in solutions of partial differential equations \cite{shivpdf} and image segmentation \cite{spie10}.

Thus, using the procedures explained in these papers, we consider a differential operator $\Delta$ depending upon the application, and find 
an integer $N$ and coefficients $c_\k^*$, $|\k|_\infty\le N$, so as the $s$--variate polynomial $P^*=\sum_{\k : |\k|_\infty \le N} c_\k^*T_\k$ minimizes
\be\label{msndef}
\int_{[-1,1]^s}|\Delta P(\t)|^2\frac{d\t}{\sqrt{(1-(t^1)^2)\cdots (1-(t^s)^2)}}
\ee
over all $s$--variate polynomials $P$ of coordinatewise degree $\le N$, subject to the conditions 
\be\label{intcond}
P(\Phi(\x_j))=f(\x_j), \qquad \x_j\in\C.
\ee 
The polynomial $P^*$ then defines an extension of $f$ to the tubular neighborhood of  the local coordinate patch of $\XX$.\\

\noindent
\textbf{Acknowledgments}\\

We would like to thank Professor Tomaso Poggio for his comments and for sharing the manuscript \cite{poggio_deep_net_2015} with us.

\newpage
\bibliographystyle{plain}
\bibliography{/Users/hrushikesh/Documents/hrushikesh/pctexfiles/hrushikesh}

\end{document}